%% file: icassp2024.tex
\documentclass{article}
\usepackage{spconf,amsmath,graphicx}
\usepackage{amsfonts}
\usepackage{multirow}
\usepackage{arydshln}
\usepackage{todonotes}
\usepackage{times}
\usepackage{microtype}
\usepackage[hidelinks]{hyperref}

\title{JPIS: A JOINT MODEL FOR PROFILE-BASED INTENT DETECTION AND SLOT FILLING WITH SLOT-TO-INTENT ATTENTION}
\name{Thinh Pham,\ \ Dat Quoc Nguyen}
\address{VinAI Research, Vietnam\\
 \{v.thinhphp1, v.datnq9\}@vinai.io}
\begin{document}
%
\maketitle
\begin{abstract}
Profile-based intent detection and slot filling are important tasks aimed at reducing the ambiguity in user utterances by leveraging user-specific supporting profile information \cite{xu2022text}. 
However, research in these two tasks has not been extensively explored. To fill this gap, we propose a joint model, namely JPIS, designed to enhance profile-based intent detection and slot filling. JPIS incorporates the supporting profile information into its encoder and introduces a slot-to-intent attention mechanism to transfer slot information representations to intent detection. 
Experimental results show that our JPIS substantially outperforms previous profile-based models, establishing a new state-of-the-art performance in overall accuracy on the Chinese benchmark dataset  ProSLU \cite{xu2022text}. 
\end{abstract}

\input{scripts/intro}

\input{scripts/model}
\input{scripts/exp}
\input{scripts/conclusion}


\bibliographystyle{IEEEbib}
\bibliography{refs}

\end{document}

%% file: scripts/intro.tex
\section{Introduction}
Recent studies on intent detection and slot filling have explored the effectiveness of joint models in enhancing overall performance, thanks to the high inherent correlation between intents and slots \cite{weld2022survey}. Some research studies \cite{goo-etal-2018-slot, li-etal-2018-self, 8907842} introduce frameworks for transferring intent information to the slot filling task, while others \cite{e-etal-2019-novel, wang-etal-2018-bi} propose models that incorporate slot representations as external knowledge in intent detection. Subsequently, numerous joint models have been proposed to leverage the dependencies between the two tasks by integrating attention mechanisms \cite{zhang-etal-2019-joint, qin-etal-2019-stack, chen2019bert, 9414110, dao21_interspeech,9747843,dgif,MISCA}.
With the advancements in deep learning and pre-trained language models \cite{devlin-etal-2019-bert, liu2019roberta}, joint intent detection and slot filling models have reached a significant overall accuracy of 92-94\% on standard benchmark datasets \cite{weld2022survey}.

Despite achieving strong performance, most existing studies are solely based on plain text, assuming that it suffices to accurately capture intents and slots. However, this assumption may not hold in many real-world situations where user utterances can be ambiguous. For instance, the utterance ``Book a ticket to Hanoi'' is ambiguous, making it challenging to correctly identify its intent, which could involve booking a plane, train or bus ticket. Therefore, relying solely on the utterances' text to predict their intent and slots may prove insufficient. 

The work in \cite{xu2022text} marks the first attempt to address this issue by introducing profile-based intent detection and slot filling tasks. These tasks take into account the user's profile information as additional knowledge to mitigate the ambiguity of the user's utterance. In this context, profile information plays a crucial role in predicting intents and slots. In the absence of profile information, even state-of-the-art models, such as those in \cite{goo-etal-2018-slot, e-etal-2019-novel, wang-etal-2018-bi, qin-etal-2019-stack}, achieve an overall accuracy of at most 44\% \cite{xu2022text}. Here, a profile-based intent detection and slot filling system can leverage two types of supporting profile information to reduce the ambiguity in utterances: User Profile and Context Awareness. The User Profile comprises a set of user-associated feature vectors representing the probability distribution of user preferences and attributes, such as transportation and audio-visual application preferences. Similarly, Context Awareness includes a list of vectors that indicate the user's state and status, including geographic location and the user's movement patterns. Furthermore, a knowledge graph might be utilized as additional information to disambiguate mentions with the same name but different entity types.

\begin{figure*}[!t]
    \centering
    \includegraphics[width=16cm]{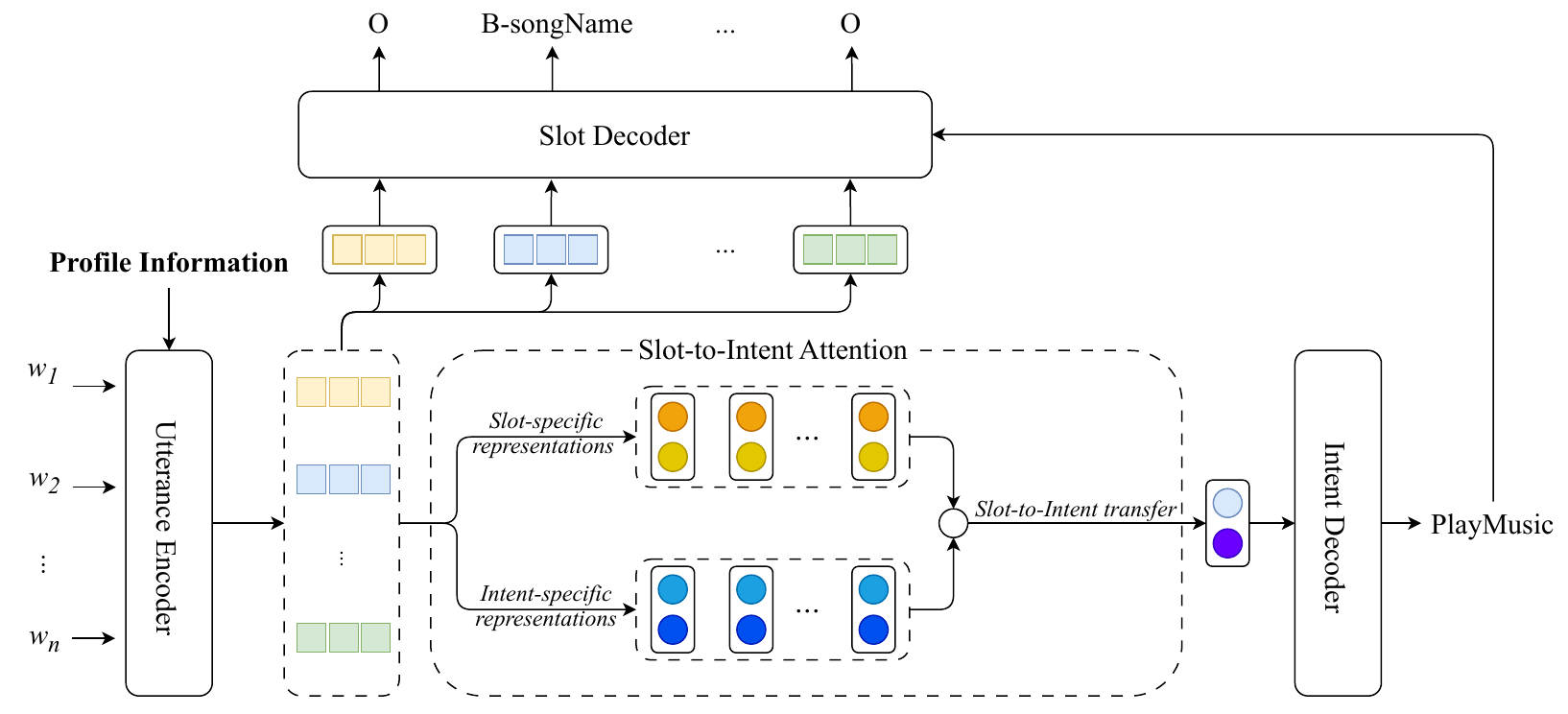}
    \caption{The architecture illustration of our model JPIS.}
    \label{fig:model}
\end{figure*}

While profile-based intent detection and slot filling are two important tasks that reflect real-world scenarios, research into these problems remains under-explored. To the best of our knowledge, the work in \cite{xu2022text} is the only one that injects supporting profile information into intent detection and slot filling models, achieving the highest overall accuracy at 82.3\%.

In this paper, we propose JPIS---a \textbf{j}oint model to further enhance the accuracy performance of \textbf{p}rofile-based \textbf{i}ntent detection and \textbf{s}lot filling.\footnote{Our JPIS implementation is available at: \url{https://github.com/VinAIResearch/JPIS} } 
JPIS follows \cite{xu2022text} to incorporate the supporting profile information into its encoder. Additionally, it introduces a slot-to-intent attention mechanism designed to facilitate the transfer of slot information into intent detection. Experiments  show that our JPIS achieves a new state-of-the-art overall accuracy at about 86.7\% on the benchmark dataset ProSLU \cite{xu2022text}. Furthermore, we conduct an ablation analysis to assess the contributions of the slot-to-intent attention mechanism and the integration of supporting profile information.

%% file: scripts/model.tex
\section{Our model JPIS}
\label{sec:model}

We formulate the profile-based intent detection and slot filling tasks as sequence classification and BIO scheme-based token classification problems, respectively. Figure \ref{fig:model} illustrates the architecture of our JPIS model, which comprises four main components: (i) Utterance Encoder, (ii) Slot-to-Intent Attention, (iii) Intent Decoder and (iv) Slot Decoder. Here, the utterance encoder incorporates the supporting profile information to generate feature vectors for word tokens in the input utterance. The slot-to-intent attention component utilizes these features to derive label-specific vector representations of intents and slot labels. It then employs slot label-specific vectors to guide intent detection, resulting in a weighted sum of intent label-specific vectors. The intent decoder component takes this weighted sum vector as input to predict the intent label of the input utterance. Finally, the slot decoder leverages the representation of the predicted intent label and word-level feature vectors from the utterance encoder to predict a slot label for each word token in the input.

\medskip 
\noindent\textbf{Utterance Encoder}: Given an input utterance with $n$ word tokens $w_1, w_2,...,w_n$, the utterance encoder first creates a vector $\mathbf{e}_{i}  \in \mathbb{R}^{d_e}$ to represent the $i$-th word token $w_i$ by  concatenating contextual word embeddings $\mathbf{e}_{i}^{\text{BiLSTM}}$ and $\mathbf{e}_{i}^{\text{SA}}$:

\begin{eqnarray}
\mathbf{e}_{i} &=& \mathbf{e}_{i}^{\text{BiLSTM}} \oplus \mathbf{e}_{i}^{\text{SA}} \label{equation:1}
\end{eqnarray}

\noindent Here, following \cite{xu2022text}, 
we feed a sequence of real-valued word embeddings $\mathbf{e}_{w_1}$, $\mathbf{e}_{w_2}$,... $\mathbf{e}_{w_n}$ into a single bi-directional LSTM layer \cite{hochreiter1997long} and a single self-attention layer \cite{vaswani2017attention} to generate the contextual vectors $\mathbf{e}_{i}^{\text{BiLSTM}}$ and $\mathbf{e}_{i}^{\text{SA}}$, respectively. 

Given that the supporting profile information for the input utterance includes $m$ user-associated feature vectors $\mathbf{x}^\mathrm{UP}_1,...,\mathbf{x}^\mathrm{UP}_m$ and $t$ context awareness vectors  $\mathbf{x}^\mathrm{CA}_1,...,\mathbf{x}^\mathrm{CA}_t$. 
The utterance encoder creates a matrix $\mathbf{P} \in \mathbb{R}^{d_p \times (m+t)}$ representing the profile information, by using projection matrices $\mathbf{W}^\mathrm{UP}_j$ and $\mathbf{W}^\mathrm{CA}_j$:

\begin{eqnarray}
    \mathbf{p}^\mathrm{UP}_j &=& \mathbf{W}^\mathrm{UP}_j\mathbf{x}^\mathrm{UP}_j \label{eqn:p_up} \\
    \mathbf{p}^\mathrm{CA}_j &=& \mathbf{W}^\mathrm{CA}_j\mathbf{x}^\mathrm{CA}_j \label{eqn:p_ca} \\
    \mathbf{P} &=& [\mathbf{p}^\mathrm{UP}_1,...,\mathbf{p}^\mathrm{UP}_m, \mathbf{p}^\mathrm{CA}_1,...,\mathbf{p}^\mathrm{CA}_t] \label{eqn:p_info}
\end{eqnarray}

To incorporate the supporting profile information into each  input word token, we also follow \cite{xu2022text} to apply the multiplicative attention mechanism  \cite{luong-etal-2015-effective}:

\begin{eqnarray}
    \alpha_{i, j} &=& \frac{\mathrm{exp}(\mathbf{e}_i\mathbf{W}^\mathrm{P}\mathbf{P}_{*,j})}{\sum_{k=1}^{m+t}\mathrm{exp}(\mathbf{e}_i\mathbf{W}^\mathrm{P}\mathbf{P}_{*,k})} \\
    \mathbf{e}_i' &=& \sum_{j=1}^{m+t}\alpha_{i, j}\mathbf{P}_{*,j}
\end{eqnarray}

\noindent where $\mathbf{W}^\mathrm{P} \in \mathbb{R}^{d_e \times d_p}$ is a weight  matrix, and $\mathbf{P}_{*,j}$ denotes the $j$-th column vector of the profile representation matrix $\mathbf{P}$. 

For the $i$-th word token, we concatenate its contextual vector $\mathbf{e}_i$ and its profile information vector $\mathbf{e}_i'$ to obtain the final vector $\mathbf{u}_i \in \mathbb{R}^{d_u}$ where $d_u = d_e + d_p$. Vectors $\mathbf{u}_i$ are concatenated to formulate an encoding matrix $\mathbf{U} \in \mathbb{R}^{d_u \times n}$ as:

\begin{eqnarray}
    \mathbf{u}_i &=& \mathbf{e}_i \oplus \mathbf{e}_i' \\
    \mathbf{U} &=& [\mathbf{u}_1, \mathbf{u}_2,...,\mathbf{u}_n] \label{eqn:u}
\end{eqnarray}

\medskip
\noindent \textbf{Slot-to-Intent Attention}: Most previous joint models consider aligning the importance of intent information to guide slot filling. However, research has demonstrated that employing slot filling could also enhance intent detection \cite{e-etal-2019-novel}. Therefore, we introduce a simple yet effective slot-to-intent attention mechanism for integrating slot  information into intent detection. 

Our slot-to-intent attention mechanism first adapts the label attention mechanism from \cite{vu2020label} to extract label-specific vector representations. Formally, we take $\mathbf{U}$ as input to compute label-specific attention weight matrices $\mathbf{A}^\mathrm{I}$ and $\mathbf{A}^\mathrm{S}$, then multiply $\mathbf{U}$ with these attention weight matrices  to obtain label-specific representation matrices $\mathbf{V}^\mathrm{I} \in \mathbb{R}^{d_u \times |\mathrm{L}^\mathrm{I}|}$ and $\mathbf{V}^\mathrm{S} \in \mathbb{R}^{d_u \times |\mathrm{L}^\mathrm{S}|}$:

\begin{eqnarray}
    \mathbf{A}^\mathrm{I} &=& \mathrm{softmax}(\mathbf{Z}^\mathrm{I} \times \mathrm{tanh}(\mathbf{Q}^\mathrm{I} \times \mathbf{U})) \\
    \mathbf{A}^\mathrm{S} &=& \mathrm{softmax}(\mathbf{Z}^\mathrm{S} \times \mathrm{tanh}(\mathbf{Q}^\mathrm{S} \times \mathbf{U})) \\
    \mathbf{V}^\mathrm{I} &=& \mathbf{U} \times (\mathbf{A}^\mathrm{I})^\top \\
    \mathbf{V}^\mathrm{S} &=& \mathbf{U} \times (\mathbf{A}^\mathrm{S})^\top
\end{eqnarray}

\noindent where $\mathrm{softmax}$ is performed at the row level; and $\mathbf{Z}^\mathrm{I} \in \mathbb{R}^{|\mathrm{L}^\mathrm{I}| \times d_a}$,  $\mathbf{Z}^\mathrm{S} \in \mathbb{R}^{|\mathrm{L}^\mathrm{S}| \times d_a}$ and $\mathbf{Q}^\mathrm{I}, \mathbf{Q}^\mathrm{S} \in \mathbb{R}^{d_a \times d_u}$ are weight matrices. Here, $\mathrm{L}^\mathrm{I}$ and $\mathrm{L}^\mathrm{S}$ are the intent label set and slot label set, respectively. 
The $j$-th column vectors $\mathbf{V}^\mathrm{I}_{*,j}$  and $\mathbf{V}^\mathrm{S}_{*,j}$ are referred to as representation vectors of the input utterance w.r.t. the $j^{\text{th}}$ label in $\mathrm{L}^\mathrm{I}$ and $\mathrm{L}^\mathrm{S}$, respectively.

The slot-to-intent attention mechanism then 
simplifies the parallel co-attention \cite{lu2016hierarchical} to calculate 
the similarity between intent and slot labels by using the label-specific representations $\mathbf{V}^\mathrm{I}$ and $\mathbf{V}^\mathrm{S}$. Specifically, it computes a bilinear attention matrix $\mathbf{C} \in \mathbb{R}^{|\mathrm{L}^\mathrm{S}| \times |\mathrm{L}^\mathrm{I}|}$ between intent and slot label types as:

\begin{equation}
    \mathbf{C} = \mathrm{tanh}((\mathbf{V}^\mathrm{S})^\top \times \mathbf{W}^\mathrm{C} \times \mathbf{V}^\mathrm{I})
\end{equation}

\noindent where $\mathbf{W}^\mathrm{C} \in \mathbb{R}^{d_u \times d_u}$ is a  weight matrix. 

After that, the mechanism allows information transfer from slot to intent by employing $\mathbf{C}$ as a feature matrix for computing an attention weight vector $\mathbf{a} \in \mathbb{R}^{|\mathrm{L}^\mathrm{I}|}$ as:

\begin{eqnarray}
    \mathbf{H} &=& \mathrm{tanh}(\mathbf{W}^\mathrm{I} \times \mathbf{V}^\mathrm{I} + (\mathbf{W}^\mathrm{S} \times \mathbf{V}^\mathrm{S}) \times \mathbf{C}) \\
    \mathbf{a} &=& \mathrm{softmax}(\mathbf{w}^\mathrm{a} \times \mathbf{H})
\end{eqnarray}

\noindent where $\mathbf{W}^\mathrm{I}, \mathbf{W}^\mathrm{S} \in \mathbb{R}^{d_c \times d_u}$ and $\mathbf{w}^\mathrm{a} \in \mathbb{R}^{d_c}$.

The final vector representation of the input utterance for intent detection is calculated as the weighted sum of the intent label-specific column vectors $\mathbf{V}^\mathrm{I}_{*,j}$:

\begin{equation}
    \mathbf{g} = \sum_{j=1}^{|\mathrm{L}^\mathrm{I}|}\mathrm{a}_j\mathbf{V}^\mathrm{I}_{*,j}
    \label{eqn:g}
\end{equation}

\medskip
\noindent \textbf{Intent Decoder}: The intent decoder takes $\mathbf{g}  \in \mathbb{R}^{d_u}$ to predict the intent label $y^\mathrm{I} = \mathrm{argmax}(\mathrm{softmax}(\mathbf{W}^\mathrm{ID}\mathbf{g}))$ for the input utterance (here, $\mathbf{W}^\mathrm{ID} \in \mathbb{R}^{|\mathrm{L}^{\mathrm{I}}|\times d_u}$). During training, a cross entropy loss $\mathcal{L}_\mathrm{ID}$ is calculated for predicting the label $y^\mathrm{I}$.

\medskip
\noindent \textbf{Slot Decoder}: To align the importance of the intent with each input token (i.e. intent-to-slot information transfer), following a common practice \cite{xu2022text, qin-etal-2019-stack}, the slot decoder also represents the predicted intent label $y^\mathrm{I}$ from the intent decoder by an embedding vector $\mathbf{e}_{y^\mathrm{I}} \in \mathbb{R}^{d_y}$. Then it concatenates each feature vector $\mathbf{u}_i$ (from Equation \ref{eqn:u}) with $\mathbf{e}_{y^\mathrm{I}}$ to create a slot filling-specific vector $\mathbf{s}_i$:

\begin{equation}
    \mathbf{s}_i = \mathbf{u}_i \oplus \mathbf{e}_{y^\mathrm{I}} \label{eqn:slot}
\end{equation}

The slot decoder projects each $\mathbf{s}_i$ into the $2|\mathrm{L}^\mathrm{S}|+1$ vector space and applies a linear-chain CRF \cite{lafferty2001conditional} to predict the corresponding slot for the $i$-th token. Here, $2|\mathrm{L}^\mathrm{S}|+1$ is the number of BIO-based slot 
tag labels (including the ``O'' label). A cross-entropy loss $\mathcal{L}_\mathrm{SF}$ is computed for slot filling during training while the Viterbi algorithm is used for inference.

\medskip
\noindent \textbf{Joint Training}: The final training objective loss $\mathcal{L}$ is a weighted sum of the intent detection and slot filling losses:
\begin{equation}
    \mathcal{L} = \lambda\mathcal{L}_\mathrm{ID} + (1-\lambda)\mathcal{L}_\mathrm{SF}
\end{equation}

%% file: scripts/exp.tex
\section{Experiments}
\label{sec:exp}
\subsection{Benchmark dataset and Evaluation metrics}

We conduct experiments on the Chinese dataset  ProSLU \cite{xu2022text}, which is the only publicly available benchmark with supporting profile information. ProSLU consists of 4196, 522 and 531 utterances for training, validation and test, respectively. Here, each utterance has 4 user-associated feature vectors and 4 context awareness vectors (i.e. $m=4$ and $t=4$).

We use standard evaluation metrics, including intent accuracy for intent detection, slot F\textsubscript{1} score for slot filling, and overall accuracy which is the percentage of utterances where both intent and slots are correctly predicted.

\subsection{Implementation details}

In the utterance encoder component, we set the dimensionality of the self-attention layer output to 128 and the dimensionality of the LSTM hidden states in the BiLSTM to 64, resulting in $d_e = 128 + 64 * 2 = 256$. We also set $d_p$ to 128, $d_a$ to 128, $d_c$ to 256, and $d_y$ to 128, thus making $d_u = d_e + d_p = 384$. 

We also experiment with another setting of utilizing pre-trained language models (PLMs), considering the representation of the first subword as the word representation. That is, following \cite{xu2022text}, $\mathbf{e}_{i}$ from Equation \ref{equation:1} is now computed as $\mathbf{e}_{i} = \mathrm{PLM}({w_{1:n}, i})$. 

We initialize the model parameters randomly and use the Adam optimizer \cite{KingmaB14} to optimize $\mathcal{L}$ with a batch size of 32 and a dropout rate of 0.4. We perform a grid search on the validation set, selecting the Adam initial learning rate from \{2e-4, 4e-4, 6e-4, 8e-4\} and the mixture weight $\lambda$ from \{0.1, 0.2, ..., 0.9\}. The model is trained for 50 epochs, and we choose the checkpoint with the highest overall accuracy on the validation set for evaluation on the test set. All the results we report are averages from 5 runs with 5 different random seeds.


\begin{table}[!t]
    \centering
        \caption{Obtained results without PLM. Note that all the baseline models have already been expanded to incorporate supporting profile information. The reported results for these baseline models are taken from \cite{xu2022text}.}
        \medskip
    \resizebox{\columnwidth}{!}{
    \setlength{\tabcolsep}{0.3em}
    \begin{tabular}{l|c|c|c}
    \hline
       \textbf{Model}  & Intent (Acc) & Slot (F\textsubscript{1}) & Overall (Acc)\\
    \hline
        SF-ID \cite{e-etal-2019-novel} & 83.24 & 73.70 & 68.36 \\
        Slot-Gated \cite{goo-etal-2018-slot} & 83.24 & 74.18 & 69.11 \\
        Bi-Model \cite{wang-etal-2018-bi} & 82.30 & 77.76 & 73.45 \\
        AGIF \cite{qin-etal-2020-agif} & 81.54 & 80.57 & 74.95 \\
        Stack-Propagation \cite{qin-etal-2019-stack}& 83.99 & 81.08 & 78.91\\
        General-SLU \cite{xu2022text} & 85.31 & 83.27 & 79.10 \\
        GL-GIN \cite{qin-etal-2021-gl} & 85.69 & 82.70 & 79.28 \\
    \hline
        Our JPIS & \textbf{87.95} & \textbf{85.76} & \textbf{82.30} \\
    \hline
    \end{tabular}}
    \label{tab:results}
\end{table}

\subsection{Main results}

\textbf{Results without PLM}:  Table \ref{tab:results} presents performance results obtained  without the use of a PLM for our JPIS model and competitive baselines on the test set.

Table \ref{tab:results} shows that our JPIS outperforms all the previous baselines across all three evaluation metrics. In particular, when compared to the previous best results, JPIS achieves substantial absolute performance improvements ranging from 2.5\% to 3.0\% in all three metrics. The most substantial improvement is observed in overall accuracy, which has increased from 79.28\% to 82.30\%. This clearly demonstrates the effectiveness of both intent-to-slot and slot-to-intent information transfer within the model architecture. It is also worth noting that the supporting profile information is utilized effectively, resulting in a positive impact on utterance representations and enhancing the interaction between intent and slot labels through the label encoder.

\begin{table}[!t]
    \centering
        \caption{Overall accuracies with PLMs. Results reported for the baseline model ``General-SLU'' are taken from  \cite{xu2022text}.}
        \medskip
    \setlength{\tabcolsep}{0.3em}
    \begin{tabular}{l|c|c}
    \hline
         \multirow{2}{*}{Models} & \multicolumn{2}{c}{Overall accuracy}  \\
         \cline{2-3}
         & General-SLU \cite{xu2022text} & Our JPIS \\
    \hline
        w/o PLM & 79.10 & \textbf{82.30} \\
    \hdashline
        w/ Chinese BERT \cite{cui-etal-2020-revisiting} & 80.98 & \textbf{85.46}  \\
        w/ Chinese RoBERTa \cite{cui-etal-2020-revisiting} & 81.73  & \textbf{86.14}  \\
        w/ Chinese  XLNet \cite{cui-etal-2020-revisiting} & 81.17  & \textbf{86.25}  \\
        w/ Chinese ELECTRA \cite{cui-etal-2020-revisiting} & 82.30 & \textbf{86.67}  \\
        
    \hline
    \end{tabular}
    \label{tab:pretrained}
\end{table}

\medskip
\noindent \textbf{State-of-the-art results with PLM}: Following \cite{xu2022text}, we also report the overall accuracy results achieved with PLMs on the test set. Table \ref{tab:pretrained} presents obtained results comparing our JPIS and the baseline ``General-SLU'' \cite{xu2022text} when combined with different PLMs. 
Unsurprisingly, the PLMs generating high-quality contextual word representations notably contribute to improving the performance of both JPIS and ``General-SLU'', resulting in overall accuracy increases of about 2\% to 4.4\%. 
Clearly, JPIS consistently outperforms ``General-SLU'' by substantial margins across all experimented PLMs, achieving absolute improvements ranging from 4.4\% to 5.1\%, establishing a new state-of-the-art overall accuracy at 86.67\%.



\begin{table}[!t]
    \centering
        \caption{Ablation results.}
        \medskip
    \resizebox{\columnwidth}{!}{
    \setlength{\tabcolsep}{0.3em}
    \begin{tabular}{l|c|c|c}
    \hline
         Model & Intent (Acc) & Slot (F\textsubscript{1}) & Overall (Acc)  \\
    \hline
        Our JPIS & \textbf{87.95} & \textbf{85.76} & \textbf{82.30} \\
    \hdashline
        \quad  w/o Slot-to-Intent & 84.97 & 83.00 & 79.89\\
        \hdashline
        \quad w/o User Profile (UP) & 50.40 & 45.78 & 46.89 \\
       \quad  w/o Context Awareness (CA)  & 80.26 & 80.71 & 75.03 \\
       \quad  w/o UP \& w/o CA & 42.22 & 39.94 & 38.79 \\
    \hline
    \end{tabular}}
    \label{tab:ablation}
\end{table}

\subsection{Ablation study}
\label{sec:ablation}
We conduct an ablation study to investigate the contributions of our model components.

\medskip
\noindent \textbf{Effect of slot-to-intent attention}: To verify the effectiveness of the slot-to-intent attention component, we conduct an experiment where we remove this component from our model (denoted by ``w/o Slot-to-Intent'' in Table \ref{tab:ablation}). We adjust the calculation of the vector representation $\mathbf{g}$ of the input utterance for intent detection, as shown in Equation \ref{eqn:g}, to the following common attention-based form: $\mathbf{g} = \mathrm{softmax}(\mathbf{w}^\mathrm{g} \times \mathbf{U}) \times \mathbf{U}^\top$. We find that removing the slot-to-intent attention results in a noticeable decrease of  3\% in intent accuracy. It also causes a reduction in slot F\textsubscript{1} by 2.8\% and an overall accuracy drop of 2.4\%. These findings provide clear evidence of the slot-to-intent attention's notable contribution by using slot-specific representations to enhance the prediction of intent labels.


\medskip
\noindent \textbf{Effect of supporting profile information}: 
We also evaluate the impact of different profile information types on the model's performance. In particular, we conduct the following experiments: (i)  without utilizing user profile (UP), i.e. adjusting Equation \ref{eqn:p_info} as $\mathbf{P} = [\mathbf{p}^\mathrm{CA}_1, \ldots, \mathbf{p}^\mathrm{CA}_t]$; (ii) without utilizing context awareness (CA), i.e. adjusting Equation \ref{eqn:p_info} as $\mathbf{P} = [\mathbf{p}^\mathrm{UP}_1, \ldots, \mathbf{p}^\mathrm{UP}_m]$;
(iii) without both UP and CA, adjusting Equation \ref{eqn:u} as $\mathbf{U} = [\mathbf{e}_1, \mathbf{e}_2, \ldots, \mathbf{e}_n]$. Our results, as shown in Table \ref{tab:ablation},  demonstrate significant decreases in all three evaluation metrics for all three ablated model settings: without UP, without CA, and without both UP and CA. Clearly, the model  incorporates the 
supporting profile information effectively.  




%% file: scripts/conclusion.tex
\section{Conclusion}

In this paper, we have introduced JPIS, a joint model for profile-based intent detection and slot filling. JPIS seamlessly integrates supporting profile information and introduces a slot-to-intent attention mechanism to facilitate knowledge transfer from slot labels to intent detection. Our experiments on the Chinese benchmark dataset ProSLU show that JPIS achieves a new state-of-the-art performance, surpassing previous models by a substantial margin.